\documentclass{article}

\usepackage{microtype}
\usepackage{graphicx}
\usepackage{subfigure}
\usepackage{booktabs} 

\usepackage{hyperref}

\usepackage[accepted]{icml2021}

\icmltitlerunning{FedDnC}

\begin{document}

\twocolumn[
\icmltitle{Weight Divergence Driven Divide-and-Conquer Approach for Optimal Federated Learning from non-IID Data}



\icmlsetsymbol{equal}{*}

\begin{icmlauthorlist}
\icmlauthor{Pravin Chandran}{intel}
\icmlauthor{Raghavendra Bhat}{intel}
\icmlauthor{Avinash Chakravarthy}{intel}
\icmlauthor{Srikanth Chandar}{intel}
\end{icmlauthorlist}

\begin{center}
Intel Technology India Pvt. Ltd, Bengaluru, KA, India\\
(pravin.chandran, raghavendra.bhat, avinash.chakravarthi,srikanth.chandar)@intel.com
\end{center}


\vskip 0.3in
]



\begin{abstract}
Federated Learning allows training of data stored in distributed devices without the need for centralizing training data, thereby maintaining data privacy. Addressing the ability to handle data heterogeneity (non-identical and independent distribution or non-IID) is a key enabler for the wider deployment of Federated Learning. In this paper, we propose a novel Divide-and-Conquer training methodology that enables the use of the popular FedAvg aggregation algorithm by overcoming the acknowledged FedAvg limitations in non-IID environments. We propose a novel use of Cosine-distance based Weight Divergence metric to determine the exact point where a Deep Learning network can be divided into class agnostic initial layers and class-specific deep layers for performing a Divide and Conquer training. We show that the methodology achieves trained model accuracy at par (and in certain cases exceeding) with numbers achieved by state-of-the-art Aggregation algorithms like FedProx, FedMA, etc. Also, we show that this methodology leads to compute and bandwidth optimizations under certain documented conditions.
\end{abstract}


\section{Introduction}
\label{Introduction}

Federated Learning has been proposed as a new learning paradigm to overcome the privacy regulations and communication overheads associated with central training \cite{b1_McMahan17}\cite{b2_Li2020}. In Federated Learning, a central server shares a global model with participating client devices and the model is trained on the local datasets available at the client device. The local dataset is never shared with the server, instead, local updates to the global model are shared with the server. The server combines the local updates from the participating clients using an Optimization (or Aggregation) Algorithm and creates a new version of the global model. This process is repeated for the required number of communication rounds until the desired convergence criteria are achieved.

Federated Learning differs significantly from traditional learning approaches in terms of optimization in a distributed setting, privacy preserving learning, and communication latency during the learning process \cite{b3_Bonawitz2019}. Optimization in Distributed setting differs from the traditional learning approach due to statistical and systems heterogeneity \cite{b1_McMahan17}. The statistical heterogeneity manifests itself in the form of non-independent and identical distribution (non-IID) of training data across participating clients. The non-IID condition arises due to a host of reasons that is specific to the local environment and usage patterns at the client. Causes for the skewed data distribution have been surveyed extensively and it has been proven that any real-world scale deployment of Federated Learning should address the challenges around non-IID data. A good example specific to the medical domain can be found in \cite{b5_Xu2020}. Several approaches have been studied to address the non-IID heterogeneity. Data Distillation which involves sharing of client data with central server \cite{b6_Zhao2018} \cite{b7_Lin2020}, Client specific local models or Personalization layers to customize the last few layers of the global model specific to the client data \cite{b8_Fallah2020} \cite{b9_Ghosh2020} \cite{b10_Hanzely2020} \cite{b11_Dinh2020} \cite{b12_Hanzely2020}, Novel optimization algorithms \cite{b1_McMahan17} \cite{b13_Sahu2020} \cite{b14_Wang2020} \cite{b15_Chen2020} etc. are some of these most researched approaches.  

Data Distillation techniques violate the strict privacy requirements. Client specific model approach results in multiple models, which does not cater to any specific requirement for a single model for deployment. In this paper, we focus on the Optimization Algorithm approach to address the non-IID challenge. While there are numerous state-of-the-art algorithms like FedProx \cite{b13_Sahu2020}, FedMA \cite{b14_Wang2020}, FedMAX \cite{b15_Chen2020} etc., these approaches are not productized in a large scale to the best of knowledge of the authors. Hence, we focus on the most widely deployed FedAvg algorithm \cite{b1_McMahan17} and investigate improving its ability to handle non-IID data to the same level as state-of-the-art algorithms like FedMA, FedProx, FedMAX, etc. 

The primary contribution of this paper is proposing a novel Divide-and-Conquer training methodology which in combination with FedAvg is able to meet state-of-the-art performance in simulated environment. Another contribution of this paper is the novel use of the Cosine Distance based Weight Divergence metric to partition the global model into class agnostic initial layers and class-specific deep layers. The two parts of the global model are trained in a mutually exclusive manner while freezing the other part. Under certain documented conditions, this approach also leads to better compute and bandwidth optimization.

The rest of the paper is organized as follows. Section II discusses the limitation with vanilla FedAvg algorithm while section III explains the Divide-and-Conquer methodology. We document the simulation environment, experiments, and results in the simulated environment in section IV establishing the state-of-the-art credentials of the approach. Finally, we conclude the paper and discuss possible future work in section V.

\section{FedAvg and its Challenges}
Federated Learning (FL) methods are designed to train over multiple devices, each holding their own data, with a central server driving the global learning objective across the entire network. The standard formulation of FL aims to find the minimizer of the overall population loss \cite{b13_Sahu2020} shown below.

\begin{figure}[ht]
\begin{center}
\centerline{\includegraphics[width=\columnwidth]{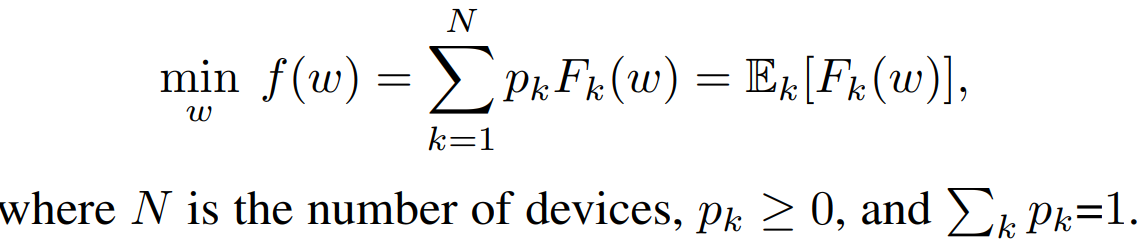}}
\label{fig:dncequate}
\end{center}
\end{figure}



In general, the local objectives measure the local empirical risk over possibly differing data distributions with samples available at each device. In a non-IID environment, the assumption of a global minimizer being representative of the overall population is not valid as every client has its own data distribution which differs from other clients and the overall population. Hence, on each client, a \textit{local objective function} based on the client's data is used as a surrogate for the global objective function. At each outer iteration, a subset of devices are selected and \textit{local solvers} are used to optimize the local objective functions of the selected client. Each client then communicates its local model updates to the central server, which aggregates them and updates the global model accordingly. In addition to the usual hyper-parameters of traditional learning like batch size, optimizer, etc., Federated Learning has additional hyper-parameters like epochs per round ($E_p$), number of communication rounds, number of participants in each round, and optimization algorithm which can be tweaked for optimal performance. 

In FedAvg, the local objective function at client $k$ is $F_k(\cdot)$, and the local solver is the stochastic gradient descent (SGD), with the same learning rate ($\eta$) and number of local epochs used on each client. At each round, a subset $K$ $<$ $N$ of the total clients are selected and run SGD locally for $E_p$ number of epochs, and then the resulting model updates are averaged. The details are summarized below.

\begin{algorithm}[tb]
   \caption{Federated Averaging Algorithm}
   \label{alg:fedavg}
\begin{algorithmic}
   \STATE {\bfseries Input:} $K$, $T$, $\eta$, $E_p$, $w^0$, $N$, $p_k$, $k=1,...,N$

    \FOR{$t=1$ {\bfseries to} $T-1$}
     \STATE Server selects a subset $S_t$ of $K$ clients at random.
     
     \STATE Server sends $w^t$ to all chosen clients.
     
     \STATE Each client $k$ $\in$ $S_t$ updates $w^t$ for $E_p$ epochs of SGD on $F_k$ with step size $\eta$ to obtain $w_k^{t+1}$
     
     \STATE Each client $k$ $\in$ $S_t$ sends $w_k^{t+1}$ back to server
     
    \STATE Server aggregates the $w$'s as $ w^{t+1} = \frac{1}{K} \sum_{k \in S_t} w_k^{t+1} $
    \ENDFOR
\end{algorithmic}
\end{algorithm}

Tuning of the hyper-parameters is a critical requirement for optimal performance of FedAvg. The number of epochs plays a critical role in convergence as more number of epochs leads to faster convergence. This comes at the cost of higher compute on client devices but with the benefit of lower communication. However, the high number of epochs has diminishing returns on the speed of convergence in non-IID conditions. For FedAvg, there is a significant drop in reduction of accuracy due to weight divergence \cite{b6_Zhao2018}. The trade-off between high number of epochs and convergence speed for FedAvg has been addressed in other optimization algorithms like FedProx, FedMA, FedMAX etc. FedProx is very similar to FedAvg but addresses the limitations of the latter by adding a proximal term to client cost functions to limit the impact of local updates within a particular range of global model. This approach allows the number of epochs to be tuned based on the non-IIDness of the client data. While it address the weight divergence issue with FedAvg, the convergence speed is slower at higher number of epochs when compared to other state-of-the-art algorithms \cite{b7_Lin2020}, \cite{b14_Wang2020} \cite{b15_Chen2020} \cite{b16_Shoham2019}. FedMA offers the best accuracy and convergence speed in comparison to others but comes with significant compute cost on the client devices. The complexity of this algorithm is also high in comparison with FedAvg or FedProx leading to restrictions on its applicability on certain NN models.

Tuning of the hyper-parameters is a critical requirement for optimal performance of FedAvg. The number of epochs plays a critical role in convergence as more number of epochs leads to faster convergence. This comes at the cost of higher compute on client devices but with the benefit of lower communication. However, the high number of epochs has diminishing returns on the speed of convergence in non-IID conditions. For FedAvg, there is a significant drop in reduction of accuracy due to weight divergence \cite{b6_Zhao2018}. The trade-off between high number of epochs and convergence speed for FedAvg has been addressed in other optimization algorithms like FedProx, FedMA, FedMAX etc. FedProx is very similar to FedAvg but addresses the limitations of the latter by adding a proximal term to client cost functions to limit the impact of local updates within a particular range of global model. This approach allows the number of epochs to be tuned based on the non-IIDness of the client data. While it address the weight divergence issue with FedAvg, the convergence speed is slower at higher number of epochs when compared to other state-of-the-art algorithms \cite{b7_Lin2020}, \cite{b14_Wang2020} \cite{b15_Chen2020} \cite{b16_Shoham2019}. FedMA offers the best accuracy and convergence speed in comparison to others but comes with significant compute cost on the client devices. The complexity of this algorithm is also high in comparison with FedAvg or FedProx leading to restrictions on its applicability on certain NN models. 

An ideal optimization algorithm should come with the simplicity and elegance of FedAvg, allow for state-of-the-art accuracy in non-IID environments with comparable or better convergence speed. In this work, we present a novel Federated Training methodology that is well suited to handle non-IID challenges using the simple FedAvg algorithm. Our methodology eliminates performance overheads associated with methods like FedMA while achieving comparable accuracy. Since FedAvg is the de-facto standard in majority production deployments, the proposed method can be easily integrated to offer significant accuracy and convergence benefits with little performance overhead.

Note on the terminology: In the rest of the document, clients will be referred to as \textit{collaborators} and the server will be referred to as {aggregator} reflecting the role they play in the overall federation.


\section{Divide-and-Conquer Training Methodology}

The impact of non-IIDness of data in Federated Learning is well researched in literature. A non-IID data environment leads to over-fitting of local models to the skewed training data at individual collaborators resulting in distortion of previously aggregated feature detectors and descent of SGD optimizer to different local minima at different collaborators.

\begin{figure}[ht]
\vskip 0.2in
\begin{center}
\centerline{\includegraphics[width=\columnwidth]{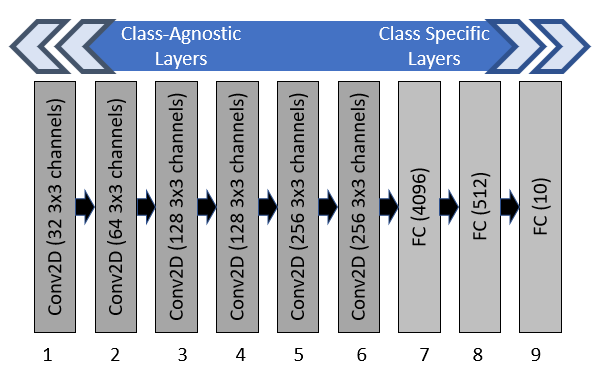}}
\caption{DNN Layer Significance - VGG9 Image Classification Topology}
\label{fig:ticktock_overview}
\end{center}
\vskip -0.2in
\end{figure}

\begin{figure}[ht]
\vskip 0.2in
\begin{center}
\centerline{\includegraphics[width=\columnwidth]{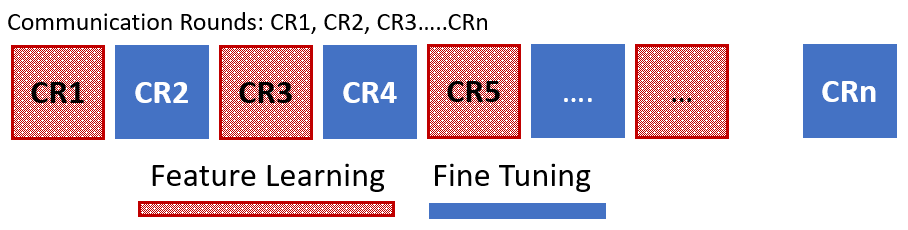}}
\caption{Divide and Conquer Training Methodology using alternate Feature-Learning and Fine-Tuning rounds. CR1, C2,. represent the communication rounds.}
\label{fig:ticktokfig}
\end{center}
\vskip -0.2in
\end{figure}

Typically, the initial layers of a Deep Neural Network (DNN) learn low level or class agnostic features and deeper layers are responsible for learning high level or class-specific features \cite{convlayers}, as illustrated for a vision architecture, VGG9 \cite{vgg_Simonyan2015},  in Figure \ref{fig:ticktock_overview}. For training paradigms like Transfer Learning \cite{tlearn1}, data scarcity mandates the use of special training methods that learn class agnostic features from generic datasets and learn class specific features for any new tasks by freezing the initial layers. This process of decoupling \textit{feature-learning} and \textit{task-learning} has been successfully applied to multiple training tasks including recent advances like Few Shot Learning \cite{b17_yan2020}. This work extends the idea to Federated Learning to address the challenges with non-IID. Our methodology involves splitting the given DNN into two parts, namely
\begin{itemize}
    \item Class Agnostic Layers
    \item Class Specific Layers
\end{itemize} 
The two parts are trained separately. Federated Learning is typically performed using several Communication Rounds (CR), where trained weights from individual collaborators are aggregated together in a central Aggregator. Our proposed method configures collaborators to perform \textit{feature-learning} and \textit{task-learning} or \textit{fine-tuning} in alternate rounds as shown in Figure \ref{fig:ticktokfig}. Weights corresponding to relevant trained layers alone are transferred over to the Aggregator, which results in communication bandwidth reduction. Communication saving is realized during model transfers in both directions (i) Transfer of global models to Collaborators and (ii) Transfer of local trained models from Collaborator to the Aggregator.

Class Agnostic layers, comprised of initial layers of the DNN architecture, are trained more aggressively as compared to Class-Specific layers. Class Agnostic layer training is treated similar to \textit{feature-learning}. Class Specific layers, consisting of deep layers are trained similar to \textit{fine-tuning}. This ensures that weight divergence across different collaborators, due to non-IIDness of constituent data is minimal and features are insulated from distortion that would otherwise occur due to combined learning of all layers.

While methods like FedProx limit weight divergence, they penalize all layers of the network and hinder learning in Class Agnostic layers. Our approach addresses this by allowing different layers of a network to train differently after grouping initial layers separately from deep layers. Training rounds are configured to alternate between \textit{feature-learning} and \textit{fine-tuning} to facilitate learning under non-IID conditions by freezing relevant layers of DNN architecture. At the beginning of a communication round, the aggregator broadcasts the desired hyper-parameter configurations to collaborators, together with specifications for layers to be frozen. The exact point at which a DNN architecture has to be broken into two parts is decided based on \textit{weight divergence} observed from the pre-pass round of training. The key contributions of our paper can be summarized to the following two key points:
\begin{itemize}
    \item Novel methodology, called \textit{Divide-and-Conquer}, to train topology in pairs of \textit{feature-learning} and \textit{fine-tuning} steps to handle non-IID conditions.
    \item Novel use of \textit{weight-divergence} metric, observed from the pre-pass round of training, to split the given DNN topology into Class Agnostic and Class Specific layers. This metric provides a measure of non-IIDness across participating collaborators as a mapping of the layers of DNN architecture they impact the most.
\end{itemize}

\begin{figure}[ht]
\vskip 0.2in
\begin{center}
\centerline{\includegraphics[width=7cm]{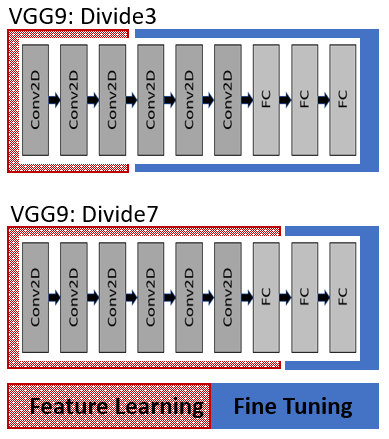}}
\caption{VGG9 : Topology Division and assignment of layers to Feature-Learning and Fine-Tuning Groups}
\label{fig:lyrsplit}
\end{center}
\vskip -0.2in
\end{figure}

Choice of layers that are chosen for base class \textit{feature-learning} as against novel class \textit{fine-tuning} is a hyper-parameter in \textit{Divide-and-Conquer} training methodology. Few options for splitting the VGG9 topology is shown in Figure \ref{fig:lyrsplit}.  For instance, Divide at layer3 assigns layers 1 to 3 for learning class agnostic features and remaining layers for learning class or task specific features. This hyper-parameter is dependent on the \textit{weight-divergence} metric which in turn reflects the non-IIDness of data.

After determining an optimal split, \textit{feature-learning} and \textit{fine-tuning} is achieved by control of other hyper-parameters like number of Epochs ($E_p$) and Learning rate ($\eta$). Fine-Tuning round of learning is scheduled using lower $E_p$ and $\eta$, which is aligned with the conditions under which FedAvg performs the best in non-IID conditions. Federated Learning at a faster pace is achieved by alternating low-level \textit{feature-learning} and high-level \textit{fine-tuning} along with appropriate hyper-parameters as described in the next section.

\section{Divide-and-Conquer: Experiments, Results, and Discussion}
This section describes the simulation environment, experiments done, and results. The comparison with other state-of-the-art approaches is also captured in the results section to establish the state-of-the-art credentials of our proposed approach.

\subsection{Experimental Setup}
We present observations from \textit{Divid-and-Conquer} on VGG9 topology using 3 different non-IID conditions as in \cite{b14_Wang2020}, which includes coverage for convolutional layers and LSTMs. Classification and NLP models used were also same as \cite{b14_Wang2020}.

\begin{itemize}
    \item Classification using Color Skewed CIFAR10 Dataset \cite{cifar}: CIFAR10 dataset is split into two groups of 5 classes each, with each class assigned uniquely to the two collaborators. To skew the data further using a 95-5\% skew pattern, 95\% of images in the first group are converted to gray-scale and 5\% of images in the second group are converted to gray-scale. This results in the first collaborator holding gray-scale dominant data and the second collaborator holding color dominant data. 
    \item Classification using Class Imbalanced CIFAR10 Data: Data is distributed non-uniformly across different collaborators to create non-IID conditions from the perspective of total training data per collaborator as well as the number of records per class.
    \item Next Character prediction model on Shakespeare dataset \cite{leaf} leveraging non-IIDness in speaking-roles: Data corresponding to each speaking-role in the play is grouped to create unique collaborators, to simulate natural non-IID condition. For the trial, we selected only clients with a minimum of 10k data points and sampled a random subset of 66 clients.
    
\end{itemize}

\subsection{Hyper-Parameter Selection}
\subsubsection{Fine Tuning Epoch and Learning Rate}
\begin{figure}[ht]
    \vskip 0.2in
    \begin{center}
    \centerline{\includegraphics[width=\columnwidth, height=5cm]{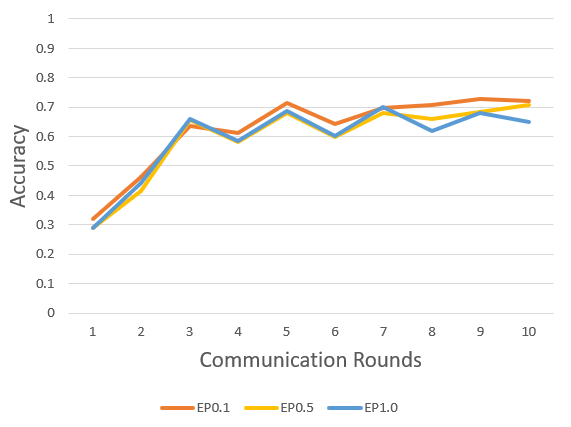}}
    \caption{Effect of Training Epochs: Ep0.1 corresponds to fine-tuning epoch that is 10\% of the value used for feature-learning} epoch
    \label{fig:epoch}
    \end{center}
    \vskip -0.2in
    \end{figure}

Divide-and-Conquer allows the use of variable hyper-parameters for different parts of the network. As discussed earlier, we train \textit{feature-learning} group more aggressively than \textit{fine-tuning} group by control of parameters like $E_p$ and $\eta$. Use of lower $E_p$ for \textit{fine-tuning} rounds results in slightly better accuracy compared to higher epochs. This is because the local models are skewed by over-fitting to non-IID data at the individual collaborators. By using lower values for epoch and learning rate for \textit{fine-tuning} rounds, we achieve better accuracy while simultaneously reducing compute requirements needed for \textit{fine-tuning} rounds. Data from Color Skewed distribution is presented in Figure \ref{fig:epoch}. This observation is in alignment with the behavior of FedAvg where a high number of $E_p$ leads to lower training accuracy due to weight divergence.\\

\subsubsection{Topology Division}
Depending on the nature and magnitude of non-IIDness, the Class Agnostic and Class Specific layers in a given model will diverge across different collaborators. We explored, weight divergence in the learned model, to guide \textit{Divide-and-Conquer (Divide-and-Conquer)} methodology. The metric given below was explored in \cite{b6_Zhao2018} .
\[
   \ W_d =  ||W_1 - W_2||/ ||W_1||
\]
where $W_d$ is Weight Divergence. 
We modified the divergence computation as below, to capture direction aware divergence to guide our \textit{Divide-and-Conquer} methodology. 
\[
   \ W_d =  Cosine Dist(W_1 , W_2)/ ||W_1||
\]

\begin{figure}[ht]
\vskip 0.2in
\begin{center}
\centerline{\includegraphics[width=\columnwidth, height=5cm]{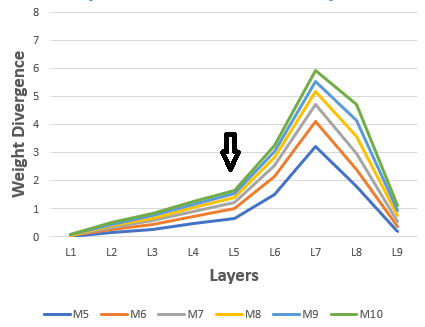}}
\caption{LayerWise Weight Divergence for Color Skewed distribution at different communication rounds. L1,L2 corresponds to layers and M5,M6 corresponds to model at end of successive communication rounds. Divergence is low for initial layers suggesting opportunities for Divide-and-Conquer}
\label{fig:wdiv1}
\end{center}
\vskip -0.2in
\end{figure}

Weight divergence from VGG9 model for Color Skewed non-IID simulation described in Section 4.1 is shown in Figure \ref{fig:wdiv1}. A pre-pass training is initially performed for 5 rounds and layer-wise divergence is computed for future rounds using the pre-pass model as a reference. Model at end of pre-pass comprising 5 rounds is M4. L1, L2 represents different layers of VGG9 while M5, M6, etc., corresponding to models from future communication rounds. Compared to prepass model M4, the divergence is low for the initial set of layers and starts to increase around Layer5. Divide-and-Conquer can be applied around this layer to split the topology for creating feature-training and fine-tuning groups.

\begin{figure}[ht]
\vskip 0.2in
\begin{center}
\centerline{\includegraphics[width=\columnwidth, height=6cm]{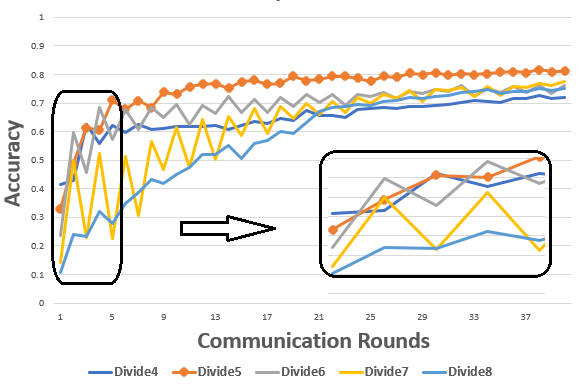}}
\caption{Training Accuracy under different Layer Division Schemes. Divide5 offers optimal results in-line with weight divergence}
\label{fig:lsplit}
\end{center}
\vskip -0.2in
\end{figure}

To validate the efficacy of $W_d$, Accuracy and convergence behavior for VGG9 under different layer division schemes were checked using a brute force sweep across different splits. Accuracy for different division schemes is presented in Figure \ref{fig:lsplit}. As discussed in Section 2, Divide5 corresponds to division after layer5.  From the figure, Divide5 offers the best accuracy and convergence speed under the given non-IID condition. All runs used 20 epochs for \textit{feature-learning} and 4 epochs for \textit{fine-tuning}. Likewise, learning rate for \textit{fine-tuning} round was half that of \textit{feature-learning}. Learning Rate Decay was also applied across the communication rounds starting from 0.001 and reducing by 10\% for every round.

For certain division schemes (ex: Divide7), large spread is seen in accuracy between \textit{feature-learning} and \textit{fine-tuning} rounds suggesting that the layer assignment strategy for the two groups is sub-optimal. For Divide7, Divide6, etc., we find that accuracy is higher for \textit{fine-tuning} rounds (CR=2,4,6,etc.,) and drops for \textit{feature-learning} rounds (CR=1,3,5,etc.). The trend however reverses for Divide5 where the accuracy is higher for \textit{feature-learning} rounds and marginally drops for \textit{fine-tuning} group. The divergence between \textit{feature-learning} and \textit{fine-tuning} is also minimal in this split.
    
When fewer layers are present in \textit{feature-learning} group as in Divide4, we find that the rate of learning starts to fall and accuracy spread between the two learning groups increases again. This suggests that Divide5 is an optimal split for this topology for this non-IID dataset thereby validating the usage of weight divergence metric to determine the point in a model where the layer split can be performed.\\

\begin{figure}[h!]
\vskip 0.2in
\begin{center}
\centerline{\includegraphics[width=\columnwidth, height=5cm]{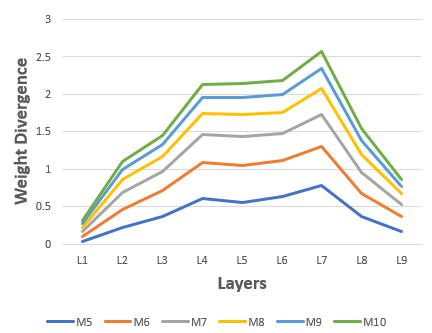}}
\caption{LayerWise Weight Divergence for class-imbalanced  distribution at different communication rounds. L1,L2 corresponds to layers and M5,M6 corresponds to model at end of successive communication rounds. Divergence is high for all layers.}
\label{fig:wdiv2}
\end{center}
\vskip -0.2in
\end{figure}

For the Class Imbalanced non-IID condition, the weight-divergence is high across all the layers of the topology [Figure \ref{fig:wdiv2}], suggesting that \textit{Divide-and-Conquer} might not offer significant benefits. Accuracy improvements was not seen in brute force sweep across different layer splits as well. For experiment sake, we chose to divide after layer8 inline with traditional \textit{fine-tuning} strategies where the last layer is used for \textit{fine-tuning}. The Next Character prediction model has 3 layers and we again use the last layer for \textit{fine-tuning}.

In the current work, layer division is determined using a pre-pass run and the scheme is fixed for the entire duration of training. Future work will extend this to explore a dynamic scheme assignment where layers from a group can be reassigned to other group based on observed trend in \textit{feature-learning} accuracy vs \textit{fine-tuning} accuracy over few communication rounds.

\subsection{Results}
Results from the \textit{Divide-and-Conquer} methodology under different non-IID scenario is presented in this section. For training we use 20 epochs for feature-learning and 4 epochs for fine-tuning. Learning rate was initialized to 0.001 and allowed to decay by 10\% for every communication round. Learning rate for fine-tuning was 50\% of learning-rate for feature-learning. 

\textit{Divide-and-Conquer} uses half the network bandwidth for data transfers compared to FedAvg, as the full model is transferred for every two communication rounds. For FedMA, results from equivalent matched averaged round is presented based on equivalency established in \cite{b14_Wang2020}. Though FedMA uses much lower communication bandwidth, compute overhead for layer matching increases with model depth as well as width, making it less desirable for practical deployments. We show that our methodology yields similar accuracy levels as more complex algorithms like FedMA in acceptable rounds of communication.\\
Note: In the tables providing the comparison across different approaches, \textit{Divide-and-Conquer} is captured under \textit{D\&C}.\\

\subsubsection{Image Classification : Color Skewed Distribution}
\begin{figure}[ht]
\vskip 0.2in
\begin{center}
\centerline{\includegraphics[width=\columnwidth]{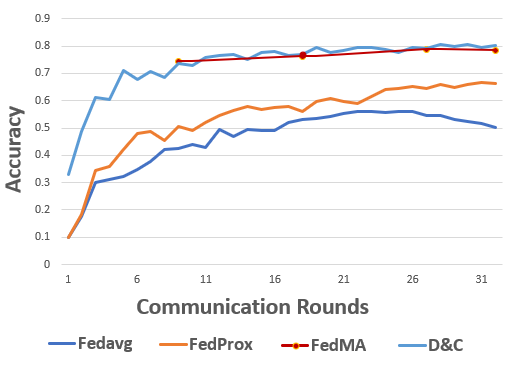}}
\caption{Accuracy Comparison for Color Skewed Distribution with 95-5\% skew. Accuracy and Convergence rate for Divide-and-Conquer (using FedAvg) is higher than FedAvg}
\label{fig:sec3sota}
\end{center}
\vskip -0.2in
\end{figure}

Training accuracy and convergence profile for different aggregation algorithms using Color Skewed 95-5\% CIFAR10 data are shown in Figure \ref{fig:sec3sota}. It can be seen that for this category of non-IIDness, the model reaches high accuracy with much smaller communication rounds compared to FedAvg. Divide5 was used for this analysis as described in the earlier section along with the same values for $E_p$ and $\eta$. Results for additional levels of Color Skew is presented in Table \ref{tab:cskew}. We chose 18 rounds of communication for the comparison to align with FedMA.

\begin{table}[h!]
\caption{Accuracy for Color Skewed Distribution for 18 communication rounds under different levels of skew for 2 collaborators. D\&C (using FedAvg) delivers high classification accuracy under this non-iidness}
\label{tab:cskew}
\vskip 0.15in
\begin{center}
\begin{small}
\begin{sc}
\begin{tabular}{lcccccr}
\toprule
\#Col & Skew & FedAvg & FedProx & FedMA & D\&C \\
\midrule
2  & 95-5\%  & 53.1\% & 56.2\% & 81.0\%  & 80.1\% \\
2  & 75-25\%  & 52.8\% & 74.6\% & 78.8\%  & 79.2\% \\
2  & 50-50\%  & 49.1\% & 67.2\% & 79.9\%  & 81.8\% \\
\bottomrule
\end{tabular}
\end{sc}
\end{small}
\end{center}
\vskip -0.1in
\end{table}

\subsubsection{Image Classification : Class Imbalance Distribution}
For Class Imbalanced data, the observed weight divergence from the pre-pass run was high for most layers. This indicates that \textit{Divide-and-Conquer} does not offer any advantages over FedAvg. As an experiment, we divided the topology at layer8 similar to traditional \textit{fine-tuning}. \textit{Divide-and-Conquer} yields slightly lower accuracy compared to FedAvg and FedMA as documented in table [\ref{tab:hetero}]. However, if bandwidth saving is not considered as a requirement and \textit{Divide-and-Conquer} is run for additional rounds to get a similar amount of model transfer as FedAvg, the performance of \textit{Divide-and-Conquer} is marginally better. This is captured in the table under the column \textit{D\&C'}. Though FedMA achieves its accuracy levels using much lower communication bandwidth, compute overhead for layer matching increases with model depth as well as width, as discussed earlier, making it less desirable for practical deployments. 

Given the results, it is clear that in cases where weight divergence suggests no clear split layer, it is recommended not to adopt \textit{Divide-and-Conquer}. As collaborator count increases, training data per collaborator decreases in the simulation environment, as the data is divided across the collaborators. This could also lead to increased divergence when \textit{feature-learning} is done aggressively on sparse data. In a truly federated set up with a large training corpus across collaborators, we expect our methodology to offer better accuracy improvements.

\begin{table}[h!]
\caption{Accuracy for Class-Imbalanced Distribution for 18 communication rounds using 5 \& 10 collaborators. Accuracy from D\&C (using FedAvg) is inline with other algorithms.}
\label{tab:hetero}
\vskip 0.15in
\begin{center}
\begin{small}
\begin{sc}
\begin{tabular}{lcccccr}
\toprule
\#Col & FedAvg & FedProx & FedMA & D\&C & D\&C'\\
\midrule
5  & 88.5\%   & 87.5\% & 87.5\% & 87.1\%  & 89.3\% \\
10  & 83.5\%  & 80.0\% & 82.5\% & 76.8\%  & 82.2\% \\

\bottomrule
\end{tabular}
\end{sc}
\end{small}
\end{center}
\vskip -0.1in
\end{table}

An extreme case of Class Imbalance based heterogeneity is when each collaborator exclusively holds data from one unique class. All the tested algorithms performed poorly (accuracy less than 15\%) under this scenario, suggesting a need for more research in this area.

\subsubsection{Next Character Prediction: speaker-role based non-IID distribution}
Results from application of \textit{Divide-and-Conquer} to a character prediction model is shown in table \ref{tab:lang}. At end of 9 communication rounds, the accuracy from \textit{Divide-and-Conquer} is comparable to other algorithms while only requiring half the amount of data transfer as FedAvg. 9 rounds of communication was chosen to align with FedMA.

\begin{table}[h!]
\caption{Accuracy for next-character prediction lstm  model for 9 communication rounds using 66 collaborators. Accuracy from D\&C (using FedAvg) is inline with other algorithms}
\label{tab:lang}
\vskip 0.15in
\begin{center}
\begin{small}
\begin{sc}
\begin{tabular}{lccccr}
\toprule
\#Col & FedAvg & FedProx & FedMA & D\&C \\
\midrule
66  & 50.8\%   & 44.6\% & 47.4\% & 49.6\% \\
\bottomrule
\end{tabular}
\end{sc}
\end{small}
\end{center}
\vskip -0.1in
\end{table}

\section{Conclusion}
In this work, we presented a weight divergence based \textit{Divide and Conquer} algorithm which builds on popular FedAvg algorithm to achieve state-of-the-art accuracy under non-IIDness. By training network in parts, our novel methodology is shown to a) Achieve faster convergence when low-level features are well-represented b) Reduce communication by half, as a consequence of training and weight exchange in parts, and c) Require less compute compared to state-of-the-art techniques like FedMA, which has performance overheads from weight matching. A static topology splitting strategy is adapted in this work, where the topology is divided at the beginning of training using a pre-pass run. Future work can explore a dynamic \textit{Divide-and-Conquer} strategy where layers are moved between \textit{feature-learning} and \textit{fine-tuning} groups based on accuracy observations during training. Future work can also explore the application of Divide-and-Conquer methodology to learning paradigms like Few Shot Learning to identify Class Agnostic layers for the backbone network.

\bibliography{main}
\bibliographystyle{icml2021}

\end{document}